\documentclass{article}

\usepackage{arxiv}

\usepackage[utf8]{inputenc} 
\usepackage[T1]{fontenc}    
\usepackage{url}            
\usepackage{graphicx}
\usepackage[numbers,sort&compress]{natbib} 
\usepackage{nicefrac}       
\usepackage{microtype}      
\usepackage{caption}

\usepackage{algorithm}
\usepackage{algpseudocode}

\usepackage{booktabs}
\usepackage{multirow}       

\usepackage{amsmath,amssymb,amsthm}
\usepackage{amsfonts}       

\usepackage{xcolor}
\usepackage{tikz}
\usetikzlibrary{arrows.meta, positioning}

\usepackage{hyperref}
\hypersetup{colorlinks=true, linkcolor=blue, citecolor=blue, urlcolor=blue}

\theoremstyle{definition}

\theoremstyle{remark}

\newcommand{\expit}{\mathrm{expit}}

\title{FaST: Factored Synthesis of Transitions for Offline Reinforcement Learning}

\renewcommand{\headeright}{A Preprint - \today}
\renewcommand{\undertitle}{A Preprint}
\renewcommand{\shorttitle}{} 

\author{
    Daniel Campa \\
    School of Mathematical and Data Sciences \\
    West Virginia University \\
    \texttt{dc00039@mix.wvu.edu} \\
    \And
    Spandan Das \\
    Department of Computer Science and Engineering \\
    Jadavpur University \\
    \texttt{spandanbhs@gmail.com} \\
    \And
    Srinjoy Das \\
    School of Mathematical and Data Sciences \\
    West Virginia University \\
    \texttt{srinjoy.das@mail.wvu.edu} \\
}

\begin{document}

\maketitle

\begin{abstract}
Synthetic data augmentation promises to scale offline reinforcement learning beyond fixed logged datasets, but leading diffusion-based generators model the transition tuple (state, action, next state, reward, terminal) as a monolith: no single factor can be natively redrawn in isolation. We present FaST (Factored Synthesis of Transitions), which factors transitions via the chain rule into state, action, dynamics, reward, and terminal stages, fitting each with an estimator matched to its statistical properties. Across twelve D4RL locomotion and maze benchmarks, TD3+BC and IQL agents trained on FaST transitions match the \textsc{SynthER} diffusion baseline while generating transitions over $90\times$ faster. The explicit action-given-state conditional can also be re-queried for a fixed state, enabling \emph{targeted action synthesis}: drawing $N$ candidate actions per generated state and keeping the highest-advantage one under a value function fit to the same data, thereby \emph{synthesizing} improved behavior during generation rather than \emph{filtering} for it post-hoc. Best-of-$10$ synthesis beats training on real data on 11 of 12 medium and medium-replay settings ($+8.9$ D4RL points on average, up to $+28.4$), and is competitive with advantage-filtered \textsc{SynthER} at a fraction of its generation cost. On expert mixtures both mechanisms degrade identically, tracing the failure to the advantage estimate itself rather than to either targeting mechanism.
\end{abstract}

\section{Introduction}

Offline reinforcement learning (RL) trains agents entirely from logged interaction, but in robotics, autonomous driving, and other high-stakes domains those logs are small, expensive to collect, and heavily skewed toward routine behavior \citep{levine2020offline}. Synthetic data augmentation offers one remedy. The standard baseline, \textsc{SynthER} \citep{lu2023synthetic}, shows that a diffusion model over the flat concatenation $(s,a,s',r,d)$ can produce transitions useful for downstream policy learning. Yet a larger synthetic dataset is not necessarily a better one: a faithful generator reproduces the behavior policy it was trained on, however mediocre~\citep{lee2024gta} (\S\ref{sec:target}). One route past that ceiling is a generator whose components can be redrawn individually. Factorizing the transition distribution for data synthesis is not itself new: model-based offline RL~\citep{yu2020mopo, kidambi2020morel} and locally-factored counterfactual methods~\citep{pitis2022mocoda} rely on partial factorizations for broader coverage, not action quality. Neither yields a reusable, policy-independent generator whose chain-rule factors are each fit by an estimator matched to the factor's statistical shape.

We adopt the chain-rule factorization
\begin{equation}
    \Pr(S,A,S',R,d) = \Pr(d\mid S,A,S',R)\cdot\Pr(R\mid S,A,S')\cdot\Pr(S'\mid S,A)\cdot\Pr(A\mid S)\cdot\Pr(S),
    \label{eq:conditional}
\end{equation}
and assign each factor an inductive bias matched to its statistical shape (\S\ref{sec:method}). The decomposition also exposes $\Pr(A \mid S)$ as an explicit conditional that can be re-queried arbitrarily often for a fixed state. Given a state-action scorer (we use the advantage of an implicit Q-learning (IQL) value function fit to $\mathcal{D}$), the generator can draw $N$ candidate actions per synthetic state and keep the best, \emph{synthesizing} transitions whose actions systematically improve on the behavior policy while the state marginal is preserved exactly and no other stage is refit. This controllability lever is effectively absent from joint generators, which admit such an intervention only through approximate workarounds (\S\ref{subsec:related}), and from model-based rollouts, whose action quality is tied to the learner's current policy.

Our contributions are two-fold:
\begin{enumerate}
    \item \textbf{FaST (Factored Synthesis of Transitions)}, a generative pipeline for offline RL transitions with stage-specific architectures. It attains D4RL scores on par with the \textsc{SynthER} diffusion baseline (Table~\ref{tab:main}) while generating transitions more than $90\times$ faster (Table~\ref{tab:throughput}).
    \item \textbf{Targeted action synthesis} as a controllability mechanism enabled by the chain-rule factorization: for each generated state we draw $N$ candidates from the explicit $\Pr(A \mid S)$ and keep the highest-advantage one. Across nine locomotion datasets and two offline-RL algorithms, training on this best-of-$10$ data beats training on real data on 11 of the 12 medium and medium-replay settings ($+8.9$ D4RL points on average). Filtering a \textsc{SynthER} pool beats best-of-$N$ on \texttt{hopper} and ties on \texttt{walker2d}, but loses on \texttt{halfcheetah}, discards state coverage, and needs a $5\times$ over-generation (Tables~\ref{tab:matrix},~\ref{tab:diffusion}). On expert mixtures both mechanisms fail identically, which traces the failure to the advantage estimate rather than the targeting mechanism (\S\ref{sec:exp-failure}).
\end{enumerate}

\section{Preliminaries}
\label{sec:preliminaries}

\subsection{Background}
\label{subsec:background}

\paragraph{Offline reinforcement learning.}
We adopt the episodic offline-RL protocol of \citet{fu2020d4rl}: a fixed dataset $\mathcal{D} = \{(s_i, a_i, s'_i, r_i, d_i)\}_{i=1}^{N_{\mathcal{D}}}$ of transitions collected by an unknown behavior policy $\pi_b$ on a Markov decision process, with $d_i \in \{0, 1\}$ the simulator-emitted termination flag. The learner seeks a policy maximizing expected discounted return and cannot further query the dynamics or $\pi_b$ ~\citep{levine2020offline}.

\paragraph{Synthetic data augmentation.}
Synthetic augmentation methods fit a generative model $\widehat{\Pr}$ of the joint transition distribution on $\mathcal{D}$, draw a synthetic dataset $\widehat{\mathcal{D}} \sim \widehat{\Pr}$, and train an offline RL algorithm on $\widehat{\mathcal{D}}$ or $\mathcal{D} \cup \widehat{\mathcal{D}}$. This decouples dataset construction from the learner. Unlike model-based offline RL, no policy-conditional rollout is required. Generators are evaluated by the downstream performance of agents trained on $\widehat{\mathcal{D}}$, typically via normalized return on benchmarks such as D4RL~\citep{fu2020d4rl}.

\paragraph{Advantage estimation from offline data.}
The targeting mechanisms of \S\ref{sec:target} score state-action pairs by an advantage estimate obtained with IQL~\citep{kostrikov2021offline}. Expectile regression fits a state value function $V$ and twin critics $Q_1, Q_2$ on $\mathcal{D}$ without querying out-of-sample actions, and we define $A_{\mathrm{IQL}}(s,a) = \min(Q_1, Q_2)(s,a) - V(s)$, so $A_{\mathrm{IQL}}(s,a) > 0$ indicates an action that beats the dataset's value baseline at $s$. The value function is fit on the same $\mathcal{D}$ as the generator.

\paragraph{Factored transition distribution.}
Equation~\ref{eq:conditional} is the chain rule for $(S,A,S',R,d)$, an identity that holds for any joint and imposes no modeling assumption of its own. The modeling content of \S\ref{sec:method} lies entirely in the per-factor estimator choice. Appendix~\ref{app:factoring} discusses the factor ordering and the distinct statistical problem each factor poses.

\subsection{Related Work}
\label{subsec:related}

\paragraph{Joint generative augmentation.} Current approaches model the joint transition distribution with deep generative models, most prominently diffusion~\citep{lu2023synthetic}. Trajectory-level variants~\citep{janner2022planning, ajay2022conditional} extend this to sequences of transitions. Joint models capture cross-factor dependence inside a single network, but pay a price on the two axes our method targets: per-sample generation cost, and component-level control. Any intervention on a single factor, such as rebiasing $\Pr(S)$ or redrawing the action for a fixed state, must either retrain the full generator on a reweighted dataset or fall back to classifier / classifier-free guidance~\citep{dhariwal2021diffusion, ho2022classifier}, inpainting-style conditional resampling~\citep{lugmayr2022repaint}, or rejection sampling, all of which yield only approximate conditionals and suffer acceptance-rate collapse in the tails. Return-guided augmentation instantiates this workaround: GTA~\citep{lee2024gta}, DiffStitch~\citep{li2024diffstitch}, and GODA~\citep{huang2025goda} all steer generation toward higher-value data, similar to our targeting, but each reaches the action indirectly rather than by resampling an explicit conditional at a fixed state. Our factorization gives $\Pr(A \mid S)$ exactly (\S\ref{sec:target}).

\paragraph{Factored data augmentation.} MoCoDA~\citep{pitis2022mocoda} exploits local causal structure \emph{within} a transition and generates counterfactuals by feeding a user-specified $(S, A)$ sampler through a learned dynamics model. We instead factor the chain rule of Eq.~\ref{eq:conditional} across the generation process and fit a specialized estimator per factor. MoCoDA's knob is a user-specified sampler over $(S, A)$. Ours is per-sample selection over the learned $\Pr(A \mid S)$ itself, which requires no auxiliary sampler and no refit: candidate actions are redrawn from the already-fitted action stage, and every other stage is reused unchanged (\S\ref{sec:target}).

\paragraph{Model-based offline RL (MBRL).} MOPO, MOReL, COMBO, RAMBO, and MOBILE~\citep{yu2020mopo, kidambi2020morel, yu2021combo, rigter2022rambo, sun2023model} learn the dynamics $\Pr(S' \mid S, A)$ on $\mathcal{D}$ and generate synthetic rollouts under the learner's policy $\pi$. The distinguishing axis is policy coupling: in MBRL rollouts are re-drawn at every learner update, so the generated data is not a reusable artifact, whereas in our case $\widehat{\mathcal{D}}$ is generated once and reused across training. For MBRL, coverage and action quality follow from $\pi$'s occupancy. In addition, MBRL caps rollouts at $H{=}1\!-\!5$ steps to bound compounding error, while our single-step generation has no rollout horizon to tune.

Appendix~\ref{app:related-work} extends this discussion to trajectory-level return-guided augmentation, to advantage-weighted regression and coverage-aware pessimism, and to the density-estimation components our pipeline composes.

\section{Method}
\label{sec:method}

\subsection{Pipeline overview}
\label{subsec:pipeline}

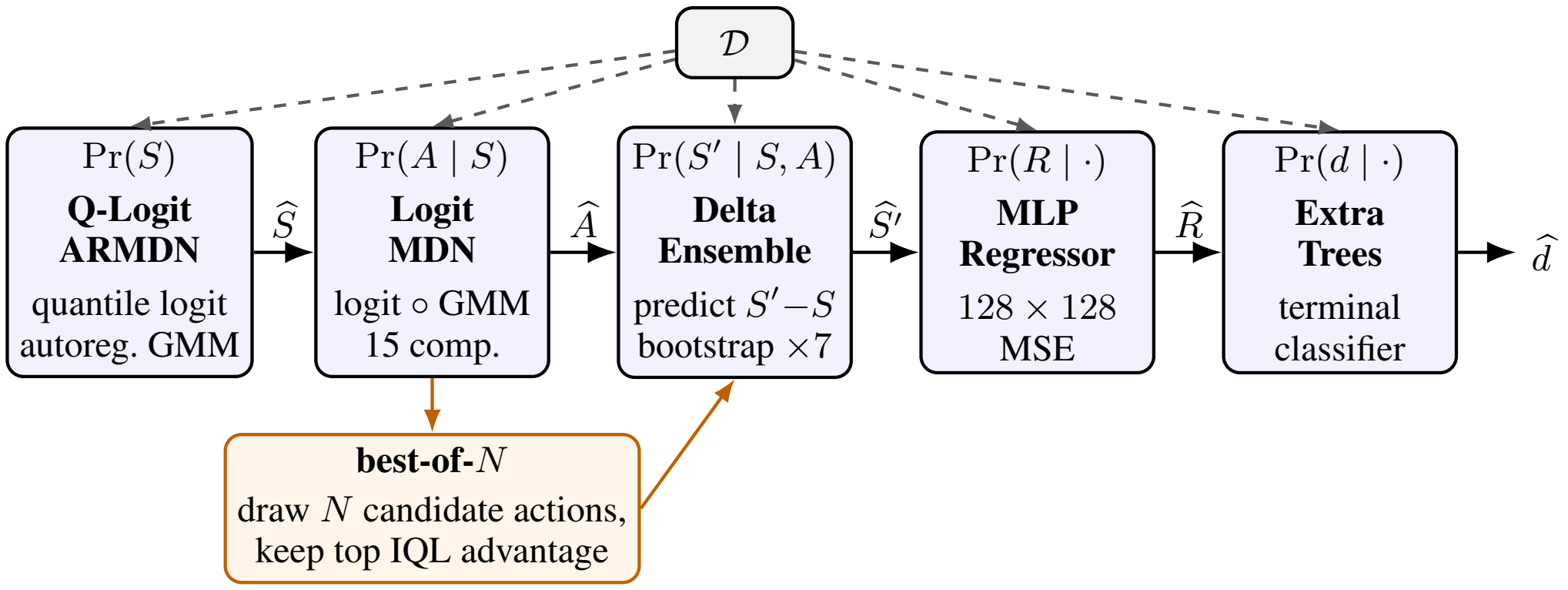
\begin{figure}[t]
\centering
\begin{tikzpicture}[
    stage/.style={draw, rounded corners, thick, minimum width=2.0cm, minimum height=1.9cm,
                  align=center, fill=blue!5, inner sep=3pt, font=\small},
    data/.style={draw, rounded corners, thick, minimum width=1.0cm, minimum height=0.6cm,
                 align=center, fill=gray!10, font=\small},
    sample/.style={-{Latex[length=2.5mm]}, thick},
    train/.style={-{Latex[length=1.8mm]}, thick, dashed, gray!70!black},
    edgelabel/.style={midway, above, font=\small},
    target/.style={draw, rounded corners, thick, minimum width=3.2cm, minimum height=0.8cm,
                   align=center, fill=orange!8, draw=orange!75!black, inner sep=3pt, font=\small},
    targetedge/.style={-{Latex[length=2mm]}, thick, orange!75!black}
]

\node[stage] (obs)  at (0.0,  0) {$\Pr(S)$\\[3pt]
    \textbf{Q-Logit}\\\textbf{ARMDN}\\[3pt]
    \small quantile  logit\\
    \small autoreg.\ GMM};

\node[stage] (act)  at (2.6,  0) {$\Pr(A\mid S)$\\[3pt]
    \textbf{Logit}\\\textbf{MDN}\\[3pt]
    \small logit $\circ$ GMM\\
    \small 15 comp.};

\node[stage] (next) at (5.2,  0) {$\Pr(S'\mid S,A)$\\[3pt]
    \textbf{Delta}\\\textbf{Ensemble}\\[3pt]
    \small predict $S'{-}S$\\
    \small bootstrap $\times 7$};

\node[stage] (rew)  at (7.8,  0) {$\Pr(R\mid \cdot)$\\[3pt]
    \textbf{MLP}\\\textbf{Regressor}\\[3pt]
    \small $128 \times 128$ \\
    \small MSE};

\node[stage] (term) at (10.4, 0) {$\Pr(d\mid \cdot)$\\[3pt]
    \textbf{Extra}\\\textbf{Trees}\\[3pt]
    \small terminal\\
    \small classifier};

\node[data] (D) at (5.2, 1.8) {$\mathcal{D}$};

\draw[train] (D) -- (obs.north);
\draw[train] (D) -- (act.north);
\draw[train] (D) -- (next.north);
\draw[train] (D) -- (rew.north);
\draw[train] (D) -- (term.north);

\draw[sample] (obs)  -- node[edgelabel]{$\widehat S$}   (act);
\draw[sample] (act)  -- node[edgelabel]{$\widehat A$}   (next);
\draw[sample] (next) -- node[edgelabel]{$\widehat{S'}$} (rew);
\draw[sample] (rew)  -- node[edgelabel]{$\widehat R$}   (term);
\draw[sample] (term.east) -- ++(0.5, 0) node[right, font=\small]{$\widehat d$};

\node[target] (bon) at (2.6, -2.2) {\textbf{best-of-$N$}\\[2pt]
    \small draw $N$ candidate actions,\\
    \small keep top IQL advantage};
\draw[targetedge] (act.south) -- (bon.north);
\draw[targetedge] (bon.east) -- (next.south);

\end{tikzpicture}
\caption{Each chain-rule factor of Eq.~\ref{eq:conditional} is fit by a stage-specialized estimator (\S\ref{subsec:state}--\S\ref{subsec:terminal}). \textbf{Dashed:} training, independent per stage on real $\mathcal{D}$. \textbf{Solid:} ancestral sampling, stage $k$ conditions on the synthetic outputs of stages $<k$. \textbf{Orange:} the optional best-of-$N$ intervention re-samples $\Pr(A\mid S)$ alone, keeping the candidate of highest IQL advantage while every other stage is reused verbatim (\S\ref{sec:target}).}
\label{fig:pipeline}
\end{figure}

Given a real transition dataset $\mathcal{D}=\{(S,A,S',R,d)\}$, we train one estimator per factor of Eq.~\ref{eq:conditional} and sample new transitions by composing them ancestrally (Figure~\ref{fig:pipeline}). Algorithm~\ref{alg:factored} states the full training and sampling procedure. The advantage estimate enters only at the candidate-selection step of \S\ref{sec:target}, and setting $N=1$ performs plain ancestral sampling. $\textsc{Fit}\,\Pr(\cdot)$ is a placeholder for the stage-specific estimators of \S\ref{subsec:state}--\S\ref{subsec:terminal}. Per-stage parameter counts and wall-clock cost are stated in Appendix~\ref{app:throughput}.

\begin{algorithm}[t]
\caption{FaST training and sampling with targeted action synthesis (best-of-$N$).}
\label{alg:factored}
\begin{algorithmic}[1]
\Require real dataset $\mathcal{D} = \{(s_i, a_i, s'_i, r_i, d_i)\}_{i=1}^{N_{\mathcal{D}}}$; candidate count $N \geq 1$
\Statex \textbf{Training (stages fit independently on $\mathcal{D}$):}
\State $\theta_S \gets \textsc{Fit}\,\Pr(S)$ on $\{s_i \in \mathcal{D}\}$
\State $\theta_A \gets \textsc{Fit}\,\Pr(A \mid S)$ on $\{(s_i, a_i) \in \mathcal{D}\}$
\State $\theta_\Delta \gets \textsc{Fit}\,\Pr(S' \mid S, A)$ on $\{(s_i, a_i, s'_i - s_i) \in \mathcal{D}\}$
\State $\theta_R \gets \textsc{Fit}\,\Pr(R \mid S, A, S')$ on $\{(s_i, a_i, s'_i, r_i) \in \mathcal{D}\}$
\State $\theta_d \gets \textsc{Fit}\,\Pr(d \mid S, A, S', R)$ on $\mathcal{D}$
\State $A_{\mathrm{IQL}} \gets$ advantage of an IQL value function fit on $\mathcal{D}$ \Comment{$\min(Q_1, Q_2) - V$}
\Statex \textbf{Sampling one synthetic transition:}
\State $\widehat S \sim p_{\theta_S}(\cdot)$
\State $\widehat A^{(1)}, \ldots, \widehat A^{(N)} \sim p_{\theta_A}(\cdot \mid \widehat S)$
\State $\widehat A \gets \arg\max_{n} A_{\mathrm{IQL}}(\widehat S, \widehat A^{(n)})$ \Comment{$N$ enters only here}
\State $\widehat{S'} \gets \,p_{\theta_\Delta}(\cdot \mid \widehat S, \widehat A)$
\State $\widehat R \gets \,p_{\theta_R}(\cdot \mid \widehat S, \widehat A, \widehat{S'})$
\State $\widehat d \gets \,p_{\theta_d}(\cdot \mid \widehat S, \widehat A, \widehat{S'}, \widehat R)$
\State \Return $(\widehat S, \widehat A, \widehat{S'}, \widehat R, \widehat d)$
\end{algorithmic}
\end{algorithm}

\paragraph{Train--sample mismatch.} All stages are trained on real inputs from $\mathcal{D}$ but, at sample time, condition on synthetic outputs of upstream stages, so the cascaded joint can diverge from the true $\Pr(S,A,S',R,d)$ even when each factor is individually accurate on real data. We mitigate this at the architecture level: the bounded-support transforms of \S\ref{subsec:state}--\S\ref{subsec:action} keep synthetic inputs on the training manifold, and the delta parametrization of \S\ref{subsec:next} bounds $\widehat{S'}$ drift relative to $\widehat S$.

\subsection{State model}
\label{subsec:state}

\paragraph{Problem setting.}
Given an offline RL dataset $\mathcal{D} = \{s_i\}_{i=1}^{N_{\mathcal{D}}}$ with $s_i \in \mathbb{R}^D$, we want synthetic observations $\hat{S}$ that reproduce two properties of $\Pr(S)$. \emph{Marginal distributions:} on the D4RL MuJoCo datasets we target, per-dimension marginals are empirically multimodal and heavy-tailed with mass concentrated at physical joint limits (Appendix~\ref{app:ablations}), so a $K$-component Gaussian mixture fit in raw space spends its components reproducing that marginal shape, leaving little capacity for cross-dimensional dependence. \emph{Cross-dimensional dependence:} the joint distribution concentrates on a low-dimensional manifold induced by the environment dynamics (joint-position/velocity correlations in mechanical systems), and violating it yields physically inconsistent states. We model $\Pr(S)$ with a mixture density network (MDN) applied in a transformed space that decouples marginal shape from dependence.

\paragraph{Quantile-logit transform.}
\label{subsubsec:qlogit}
Unlike affine z-score normalization, which preserves each dimension's multimodal or heavy-tailed shape, the per-dimension quantile--logit transform maps every marginal to a common logistic form. For each dimension $j \in \{0,\ldots,D-1\}$, the forward transform is
\begin{equation}
    T_j(s_j) \;=\; \mathrm{logit}\!\left(\hat F_j(s_j)\right), \qquad \mathrm{logit}(u) \;=\; \log\frac{u}{1-u},
    \label{eq:qlogit-fwd}
\end{equation}
where $\hat F_j$ is the rank-based empirical CDF, $\hat F_j(s_{i,j}) = \mathrm{rank}(s_{i,j}) / (N_{\mathcal{D}}+1)$, clipped to $[\epsilon, 1-\epsilon]$ with $\epsilon = 10^{-4}$ before the logit to prevent boundary blow-up. The inverse used at sample time, $T_j^{-1}(z_j) = \hat F_j^{-1}(\expit(z_j))$ with $\expit = \mathrm{logit}^{-1}$, computes $\hat F_j^{-1}$ by linear interpolation through the sorted training values.

The transform has two properties relevant to our use. (i) \emph{In-support sampling:} because $\hat F_j^{-1}$ maps into the observed range $[\min_i s_{i,j},\, \max_i s_{i,j}]$, synthetic observations $\hat S = T^{-1}(Z)$ never fall outside training support, at the cost of forgoing explicit extrapolation beyond $\mathcal{D}$ (rationale in Appendix~\ref{app:ablations}). (ii) \emph{Copula estimation:} $\expit(Z)$ has uniform marginals by construction, so $\Pr(Z)$ identifies the copula of $\Pr(S)$ \citep{nelsen2006introduction,joe2014dependence}, lifted onto $\mathbb{R}^D$ where Gaussian-mixture estimators are well-posed.

\paragraph{Architecture.}
Let $Z = T(S)$ denote the state in the transformed space. Following the real-valued autoregressive density-estimation tradition of RNADE~\citep{uria2013rnade}, we factor the joint distribution autoregressively over $D$ dimensions,
\begin{equation}
    \Pr(Z) \;=\; \Pr(Z_0)\,\prod_{j=1}^{D-1} \Pr(Z_j \mid Z_{<j}).
    \label{eq:ar-factor}
\end{equation}
$Z_0$ has a standard logistic marginal by construction and is sampled in closed form. This autoregressive MDN (ARMDN) learns only the $D{-}1$ conditionals, each modeled by a small two-hidden-layer MLP with weights $\theta_j$ that outputs the parameters of a univariate $K$-component Gaussian mixture:
\begin{equation}
    p_{\theta_j}\!\left(z_j \mid z_{<j}\right) \;=\;
    \sum_{k=1}^{K} \pi_{j,k}(z_{<j})\,\mathcal{N}\!\left(z_j;\,\mu_{j,k}(z_{<j}),\,\sigma_{j,k}^2(z_{<j})\right),
    \label{eq:state-mdn}
\end{equation}
for $j \geq 1$. Each conditional is thus a univariate Gaussian mixture, as in RNADE. Unlike RNADE, which ties a single weight-shared network across all conditionals, we fit each $\theta_j$ \emph{independently} by maximum likelihood, with its own MLP, training schedule, and early stopping on a held-out validation split. This trades RNADE's parameter efficiency for per-dimension capacity. At sample time we draw $\hat Z_0 \sim \mathrm{Logistic}(0,1)$ directly, $\hat Z_j \sim p_{\theta_j}(\cdot \mid \hat Z_{<j})$ for $j\geq 1$, and invert, $\hat S = T^{-1}(\hat Z)$. We use $K=15$ components for every dataset. Dimension ordering is fixed to the MuJoCo default column order: positions precede their corresponding velocities, which lets each velocity conditional exploit the known kinematic relationship with the already-sampled positional coordinate.

\subsection{Action model}
\label{subsec:action}

\paragraph{Problem setting.}
Given state-action pairs $\{(s_i, a_i)\}_{i=1}^{N_{\mathcal{D}}}$ with $a_i \in [-1, 1]^d$, we want to estimate $\Pr(A \mid S)$ and sample $\hat A \sim \Pr(A \mid \hat S)$ that reproduce two properties of the conditional action distribution: \emph{multimodality} (the same state can admit multiple valid actions, e.g.\ a hopper mid-fall recovering via different joint torques, so a unimodal regressor collapses to the conditional mean) and \emph{bounded support} (MuJoCo actuators impose $A \in [-1, 1]^d$). A raw-space Gaussian MDN respects multimodality but assigns mass outside $[-1, 1]^d$, so we reparameterize to bounded support.

\paragraph{Bounded-action logit transform.}
\label{subsubsec:action-logit}
Locomotion datasets concentrate many action dimensions at the actuator limits $\pm 1$, so a Gaussian mixture trained in raw action space must either leak probability mass outside $[-1, 1]^d$ or use pathologically narrow components near the boundary. We address this with a per-dimension bijection from $(-1, 1)$ to $\mathbb{R}$,
\begin{equation}
    T_j(a_j) \;=\; \log\frac{1 + a_j}{1 - a_j} \;=\; 2\,\mathrm{arctanh}(a_j),
    \qquad
    T_j^{-1}(z_j) \;=\; \tanh(z_j / 2),
    \label{eq:action-logit}
\end{equation}
applied after clipping actions to $[-1{+}\epsilon,\, 1{-}\epsilon]$ (we use $\epsilon = 10^{-3}$) so exact-boundary samples do not map to $\pm\infty$. Under $T$, boundary-saturated modes at $\pm 1$ become finite distributional modes at $\pm T(1-\epsilon) \approx \pm 7.6$, giving the Gaussian mixture an unbounded target. The inverse $\tanh(\cdot/2)$ is the same squashing used in standard bounded-policy RL parameterizations~\citep{haarnoja2018soft}, so synthetic actions land in $(-1, 1)^d$ by construction. Appendix~\ref{app:action-details} contrasts $T$ with the data-dependent quantile--logit transform of \S\ref{subsubsec:qlogit}.

\paragraph{Architecture.}
We model $\Pr(A \mid S)$ via a MDN on the logit-transformed action $Z = T(A)$. Action dimensionality is small in our benchmarks ($d \leq 6$), so we use a joint diagonal Gaussian mixture rather than an autoregressive factorization:
\begin{equation}
    \Pr(Z \mid S) \;=\; \sum_{k=1}^{K} \pi_k(S)\,\mathcal{N}\!\left(Z;\,\mu_k(S),\,\mathrm{diag}(\sigma_k^2(S))\right),
    \label{eq:action-mdn}
\end{equation}
where $(\pi_k, \mu_k, \sigma_k)_{k=1}^{K}$ are the outputs of a residual MLP $f_{\theta_A}(S)$ with sigmoid linear unit (SiLU) activations and LayerNorm. We train $\theta_A$ by minimizing the negative log-likelihood of $Z = T(A)$ on real $(S, A)$ pairs with early stopping on a held-out validation split. At sample time we condition on synthetic states, draw $k \sim \pi(\hat S)$, sample $Z$ from the $k$-th Gaussian, and output $\hat A = T^{-1}(Z)$. We use $K=15$ components throughout. Because the logit transform renders each marginal near-Gaussian, fit is insensitive to $K$ above ${\sim}3$ (Appendix~\ref{app:action-details}), so we fix $K$ generously rather than tuning it. Full settings are listed in Appendix~\ref{app:hyperparams}.

\subsection{Next-state model}
\label{subsec:next}

\paragraph{Problem setting.}
\label{subsubsec:problem-delta}
Given transition tuples $\{(s_i, a_i, s'_i)\}_{i=1}^{N_{\mathcal{D}}}$ with $s_i, s'_i \in \mathbb{R}^D$ and $a_i \in [-1,1]^d$, this stage models the conditional $\Pr(S' \mid S, A)$, extending ancestral sampling from \S\ref{subsec:state}--\S\ref{subsec:action} so that next-state draws can be conditioned on synthetic $(\widehat{S}, \widehat{A})$.

The next-state conditional differs in kind from $\Pr(S)$ and $\Pr(A \mid S)$: in the D4RL locomotion suites the simulator is near-deterministic and residual variation comes from exploration noise, so $\Pr(S' \mid S, A)$ is approximately unimodal with small variance. We further exploit temporal proximity at typical control frequencies ($s' \approx s$) by reparameterizing in residual space, $\delta = S' - S$, so a regressor does not waste capacity relearning the identity component of the map. The conditional $\delta \mid (s, a)$ is tightly concentrated near zero and approximately unimodal (Appendix~\ref{app:delta-diagnostics}), so a point predictor under mean squared error (MSE) in raw $\delta$-space captures the bulk of the conditional.

\paragraph{Architecture.}
\label{subsubsec:delta-arch}
We model $\delta = S' - S$ with a bootstrap ensemble $\{f_{\theta_{\Delta,m}}\}_{m=1}^M$ of $M$ independently-trained regressors. Each member is a residual MLP: an input projection (Linear $\to$ SiLU $\to$ LayerNorm), a stack of pre-norm residual blocks with dropout, and a linear output head to $\mathbb{R}^D$. Member $m$ is trained with MSE on $\delta$ over a bootstrap resample of the training set, using non-terminal transitions only (terminal rows pair a pre-end state with a simulator-reset $s'$ and would corrupt the conditional mean); diversity comes from per-member bootstrap, independent initialization, and dropout. Given synthetic conditioning $(\widehat{S}, \widehat{A})$, we emit $\widehat{S}' = \widehat{S} + \frac{1}{M}\sum_{m} f_{\theta_{\Delta,m}}(\widehat{S}, \widehat{A})$, averaging rather than sampling a random member. Averaging reduces prediction variance (Appendix~\ref{app:delta-diagnostics}) at no cost to fidelity, since the conditional is near-deterministic. The values of $M$, hidden widths, and dropout are given in Appendix~\ref{app:hyperparams}.

\subsection{Reward and terminal models}
\label{subsec:reward}

\paragraph{Reward.}
The reward is a near-deterministic function of $(s,a,s')$ with a unimodal, small-variance conditional (Appendix~\ref{app:reward-diagnostics}), so MSE on the conditional mean is the appropriate objective and we train directly in raw reward space. We reuse a single member ($M{=}1$) of the residual-MLP architecture of \S\ref{subsec:next} with a scalar output head, with a held-out validation split for early stopping. At sample time we emit $\widehat R = f_{\theta_R}(\widehat S, \widehat A, \widehat{S'})$.

\paragraph{Terminals.}
\label{subsec:terminal}
In the D4RL locomotion suite terminations are a deterministic function of $s'$: the simulator raises $d = 1$ when state coordinates violate axis-aligned thresholds (e.g.\ Hopper terminates when torso height or body angle leaves a fixed range), so the decision boundary is piecewise-axis-aligned and non-smooth, and the positive class is rare ($< 1\%$ of transitions on medium-replay). We retain $r$ in the conditioning set for consistency with Eq.~\ref{eq:conditional} since its dependence is near-vacuous. Both properties motivate an Extremely Randomized Trees classifier~\citep{geurts2006extremely} ($M=50$ trees) on features $(S, A, S', R)$: randomized axis-aligned splits match the threshold structure with far less capacity than a smooth-boundary regressor, and they yield ensemble diversity without bootstrap resampling, which can omit the rare positive class entirely. At sample time we emit $d$ deterministically, thresholding $p_{\theta_d}(1 \mid \cdot)$ at the cut $\tau$ that reproduces the empirical terminal rate of $\mathcal{D}$.

\subsection{Advantage-targeted generation}
\label{sec:target}

\paragraph{Motivation.}\label{subsubsec:target-motivation}
Logged offline-RL datasets are dominated by mediocre behavior, and a faithful generator reproduces this behavior. In applications such as robotics, autonomous driving, and recommendation systems, useful synthetic data is enriched with \emph{good} behavior: transitions whose actions beat the behavior policy's value baseline, as measured by the IQL advantage $A_{\mathrm{IQL}}(s,a)$ of \S\ref{subsec:background}. We consider two mechanisms.

\paragraph{Enrichment by filtering (generator-agnostic).}\label{subsubsec:target-filter}
Generate a pool of transitions, score each with $A_{\mathrm{IQL}}(\widehat S, \widehat A)$, and keep the top fraction $\rho$. We use a hard top-$20\%$ of a $5\times$ over-generated pool, so the filtered dataset is \emph{size-matched} to the untargeted one and any difference is attributable to selection, not dataset size. Filtering operates on the generator's \emph{output}, so it applies to any generator, including joint diffusion, and is our generator-agnostic baseline. Its cost is coverage: discarding $1 - \rho$ of the pool discards the states along with the actions.

\paragraph{Targeted action synthesis (best-of-$N$).}\label{subsubsec:target-bon}
Because FaST represents $\Pr(A \mid S)$ explicitly, we can instead intervene on the action factor alone: for each synthetic state $\widehat S$, draw $N$ candidate actions $\widehat A^{(1)}, \ldots, \widehat A^{(N)} \sim p_{\theta_A}(\cdot \mid \widehat S)$, keep $\widehat A = \arg\max_n A_{\mathrm{IQL}}(\widehat S, \widehat A^{(n)})$, and continue ancestral sampling through the delta, reward, and terminal stages as usual (Figure~\ref{fig:pipeline}). This \emph{synthesizes} good behavior rather than filtering for it: the state marginal, and therefore state coverage, is preserved exactly, and no stage is refit. A joint generator admits this intervention only through the approximate workarounds of \S\ref{subsec:related}, which forfeit exact draws from the learned $\Pr(A \mid S)$: guidance samples a tilted distribution rather than the conditional itself, and pushing it toward the high-advantage tail that targeting explicitly seeks degrades dynamic plausibility~\citep{lee2024gta}. Best-of-$N$ relies only on cheap over-generation ($N$ action draws per transition), which the throughput advantage of Table~\ref{tab:throughput} makes essentially free.

\section{Experiments}
\label{sec:experiments}

\subsection{Experimental setup}
\label{sec:exp-setup}

\paragraph{Benchmarks.}
We evaluate on twelve D4RL \citep{fu2020d4rl} datasets: MuJoCo locomotion (\texttt{halfcheetah}, \texttt{hopper}, \texttt{walker2d}, each at medium, medium-replay, medium-expert) and \texttt{maze2d} at three layouts (umaze, medium, large).

\paragraph{Baselines.}
We compare five sources of training data. \emph{Real} is the D4RL dataset $\mathcal{D}$; \emph{\textsc{SynthER}} is the joint-diffusion generator of \citet{lu2023synthetic}; \emph{FaST} is our pipeline of \S\ref{sec:method} without advantage targeting ($N=1$); \emph{filter} is the size-matched top-$20\%$ filter of \S\ref{sec:target}; and \emph{best-of-10} is targeted action synthesis with $N=10$.

\paragraph{Protocol.}
All results (Tables~\ref{tab:main}, \ref{tab:matrix}, and~\ref{tab:diffusion}) use one uniform protocol. For each (dataset, generator, generation seed) we draw $\widehat{\mathcal{D}}$ of $10^6$ transitions and train the agent on $\widehat{\mathcal{D}}$ alone for $10^6$ steps, evaluating every 25k steps over 10 episodes. A run's score is the mean D4RL-normalized return over the final 500k steps (the last 20 of 40 evaluation points), following standard practice \citep{fujimoto2021minimalist, tarasov2023revisiting}. We use TD3+BC \citep{fujimoto2021minimalist} and IQL \citep{kostrikov2021offline} with default hyperparameters. Synthetic arms use 3 generation seeds $\times$ 3 training seeds, and ``$\pm$'' is the std across the means of 3 generation seeds (generation variance). The real-data baseline has no generation seed, so its ``$\pm$'' is the std across 3 training seeds.

\subsection{Main results}

Table~\ref{tab:main} reports D4RL-normalized return for agents trained on each generator's untargeted synthetic dataset. FaST is on par with \textsc{SynthER} across the suite, with a mean difference of $+1.7$ in FaST's favor over the first 18 locomotion cells and most cells agreeing within a few points. The two largest gaps run in opposite directions (\texttt{hopper-medium-expert} under IQL, $101.5$ vs.\ $85.7$; \texttt{hopper-medium-replay} under TD3+BC, $52.9$ vs.\ $62.1$). On the maze2d rows FaST matches or exceeds \textsc{SynthER} on all three layouts under IQL and on two of three under TD3+BC.

\begin{table}[t]
\centering
{\footnotesize
\setlength{\tabcolsep}{4pt}
\begin{tabular}{llccc|ccc}
\toprule
& & \multicolumn{3}{c|}{\textbf{TD3+BC}} & \multicolumn{3}{c}{\textbf{IQL}} \\
\cmidrule(lr){3-5} \cmidrule(lr){6-8}
\textbf{Environment} & \textbf{Dataset} & Real & \textsc{SynthER} & FaST & Real & \textsc{SynthER} & FaST \\
\midrule
\multirow{3}{*}{halfcheetah-v2}
    & medium        & 48.2$\pm$0.1 & 48.2$\pm$0.1 & 48.0$\pm$0.2  & 48.1$\pm$0.1 & 48.8$\pm$0.1 & 48.1$\pm$0.2  \\
    & medium-replay & 44.5$\pm$0.2 & 45.2$\pm$0.2 & 44.5$\pm$0.3  & 44.0$\pm$0.1 & 46.2$\pm$0.1 & 45.3$\pm$0.5  \\
    & medium-expert & 88.6$\pm$1.6 & 77.0$\pm$2.1 & 82.0$\pm$4.3  & 93.1$\pm$1.1 & 91.3$\pm$0.7 & 92.9$\pm$0.6  \\
\midrule
\multirow{3}{*}{hopper-v2}
    & medium        & 58.5$\pm$0.7 & 61.3$\pm$0.3 & 60.7$\pm$0.5  & 63.0$\pm$1.6 & 68.3$\pm$1.0 & 67.4$\pm$4.0  \\
    & medium-replay & 62.4$\pm$6.1 & 62.1$\pm$3.2 & 52.9$\pm$11.5 & 86.9$\pm$4.0 & 97.4$\pm$3.1 & 96.5$\pm$3.0  \\
    & medium-expert & 99.1$\pm$3.8 & 94.7$\pm$0.4 & 102.4$\pm$1.4 & 79.9$\pm$5.8 & 85.7$\pm$7.1 & 101.5$\pm$3.5 \\
\midrule
\multirow{3}{*}{walker2d-v2}
    & medium        & 83.9$\pm$0.4  & 78.7$\pm$6.9  & 83.7$\pm$0.1  & 78.6$\pm$2.8  & 69.3$\pm$19.1 & 75.4$\pm$3.6  \\
    & medium-replay & 83.3$\pm$1.8  & 84.7$\pm$2.5  & 79.1$\pm$0.9  & 70.9$\pm$6.8  & 68.6$\pm$2.5  & 77.0$\pm$1.4  \\
    & medium-expert & 110.4$\pm$0.0 & 110.2$\pm$0.3 & 110.0$\pm$0.3 & 111.6$\pm$0.5 & 110.1$\pm$3.1 & 111.4$\pm$0.8 \\
\midrule
\multirow{3}{*}{maze2d-v1}
    & umaze  & 35.0$\pm$18.1  & 36.5$\pm$18.2  & 40.1$\pm$13.2 & 39.7$\pm$1.7 & 37.1$\pm$3.4 & 46.1$\pm$6.9  \\
    & medium & 43.2$\pm$33.3  & 62.0$\pm$43.8  & 62.6$\pm$39.2 & 35.4$\pm$2.7 & 33.9$\pm$1.5 & 41.0$\pm$14.6 \\
    & large  & 110.9$\pm$48.6 & 104.2$\pm$40.2 & 93.3$\pm$43.8 & 59.9$\pm$7.1 & 59.7$\pm$3.0 & 67.8$\pm$6.8  \\
\bottomrule
\end{tabular}}
\caption{D4RL normalized scores for offline RL agents trained on purely synthetic data from each generator. Evaluation with uniform protocol and mean $\pm$ std as described in \S\ref{sec:exp-setup}.}
\label{tab:main}
\end{table}

\subsection{Advantage-targeted generation}
\label{sec:exp-target}

Table~\ref{tab:matrix} evaluates both targeting mechanisms of \S\ref{sec:target} across the full locomotion suite and both learners. Generators and IQL value functions are fit on the full real dataset. A dose-response sweep over $N$ is discussed in Appendix~\ref{app:target-details}.

\begin{table}[t]
\centering
{\footnotesize
\setlength{\tabcolsep}{2.5pt}
\begin{tabular}{llcccc|cccc}
\toprule
& & \multicolumn{4}{c|}{\textbf{TD3+BC}} & \multicolumn{4}{c}{\textbf{IQL}} \\
\cmidrule(lr){3-6} \cmidrule(lr){7-10}
\textbf{Environment} & \textbf{Dataset} & Real & FaST & $+$filter & $+$BoN-10 & Real & FaST & $+$filter & $+$BoN-10 \\
\midrule
\multirow{3}{*}{halfcheetah-v2}
    & medium        & 48.2$\pm$0.1 & 48.0$\pm$0.2  & 50.4$\pm$0.3  & 53.0$\pm$0.3  & 48.1$\pm$0.1 & 48.1$\pm$0.2  & 50.4$\pm$0.0 & 52.5$\pm$0.1  \\
    & medium-replay & 44.5$\pm$0.2 & 44.5$\pm$0.3  & 47.1$\pm$0.1  & 49.5$\pm$0.2  & 44.0$\pm$0.1 & 45.3$\pm$0.5  & 47.4$\pm$0.2 & 49.9$\pm$0.2  \\
    & medium-expert & 88.6$\pm$1.6 & 82.0$\pm$4.3  & 67.0$\pm$1.8  & 65.8$\pm$6.4  & 93.1$\pm$1.1 & 92.9$\pm$0.6  & 70.2$\pm$3.2 & 71.2$\pm$2.3  \\
\midrule
\multirow{3}{*}{hopper-v2}
    & medium        & 58.5$\pm$0.7 & 60.7$\pm$0.5  & 70.2$\pm$2.1  & 77.0$\pm$1.7  & 63.0$\pm$1.6 & 67.4$\pm$4.0  & 68.3$\pm$2.4 & 71.6$\pm$3.7  \\
    & medium-replay & 62.4$\pm$6.1 & 52.9$\pm$11.5 & 87.2$\pm$2.9  & 90.8$\pm$8.8  & 86.9$\pm$4.0 & 96.5$\pm$3.0  & 77.0$\pm$6.4 & 93.8$\pm$8.0  \\
    & medium-expert & 99.1$\pm$3.8 & 102.4$\pm$1.4 & 109.1$\pm$0.5 & 102.2$\pm$4.0 & 79.9$\pm$5.8 & 101.5$\pm$3.5 & 48.9$\pm$1.2 & 44.2$\pm$10.8 \\
\midrule
\multirow{3}{*}{walker2d-v2}
    & medium        & 83.9$\pm$0.4  & 83.7$\pm$0.1  & 88.3$\pm$0.3  & 87.8$\pm$0.3  & 78.6$\pm$2.8  & 75.4$\pm$3.6  & 80.1$\pm$1.1  & 77.7$\pm$1.4  \\
    & medium-replay & 83.3$\pm$1.8  & 79.1$\pm$0.9  & 83.3$\pm$1.3  & 90.7$\pm$1.0  & 70.9$\pm$6.8  & 77.0$\pm$1.4  & 73.6$\pm$3.3  & 85.3$\pm$1.8  \\
    & medium-expert & 110.4$\pm$0.0 & 110.0$\pm$0.3 & 111.8$\pm$0.7 & 112.6$\pm$0.2 & 111.6$\pm$0.5 & 111.4$\pm$0.8 & 112.6$\pm$2.5 & 113.9$\pm$0.1 \\
\bottomrule
\end{tabular}}
\caption{Advantage-targeted generation across the locomotion suite (uniform protocol of \S\ref{sec:exp-setup}). \emph{Real}: the D4RL dataset $\mathcal{D}$. \emph{FaST}: the untargeted pipeline ($N{=}1$). \emph{$+$filter}: top-$20\%$ of a $5\times$ FaST pool (size-matched). \emph{$+$BoN-10}: targeted action synthesis with $N{=}10$. Both targeted arms build on FaST and use the same advantage estimate.}
\label{tab:matrix}
\end{table}

Table~\ref{tab:matrix} supports two findings on the medium and medium-replay tiers. The medium-expert tier, where targeting can fail badly and both mechanisms fail \emph{together}, is deferred to \S\ref{sec:exp-failure}.

First, targeted action synthesis is uniformly effective on medium and medium-replay: best-of-10 beats training on real data on 11 of those 12 settings (mean $+6.6$ on medium, $+11.3$ on medium-replay). The clearest case is \texttt{halfcheetah-medium}, where every generator in Table~\ref{tab:main} sits at the $\sim$48 ceiling that best-of-10 breaks ($53.0$/$52.5$).

Second, size-matched filtering gains too, but half as much (mean $+4.2$ on the same 12 settings), and best-of-$N$ is the stronger mechanism on 10 of the 12. The gap is widest under IQL on medium-replay, where filtering fails to beat the untargeted arm on two of the three datasets. Best-of-$N$ optimizes the action factor without discarding states, retains full coverage, and wins the same cells outright ($85.3$ on walker2d, $+14.4$ over real data).

\paragraph{Comparison to \textsc{SynthER}.}
Table~\ref{tab:diffusion} pits the FaST arms against the same interventions applied to \textsc{SynthER} across the three medium-replay datasets.

\begin{table}[t]
\centering
{\footnotesize
\setlength{\tabcolsep}{4pt}
\begin{tabular}{llccc|ccc}
\toprule
& & \multicolumn{3}{c|}{\textbf{TD3+BC}} & \multicolumn{3}{c}{\textbf{IQL}} \\
\cmidrule(lr){3-5} \cmidrule(lr){6-8}
\textbf{Dataset} & \textbf{Data source} & Untargeted & $+$filter & $+$BoN-10 & Untargeted & $+$filter & $+$BoN-10 \\
\midrule
\multirow{2}{*}{halfcheetah-mr}
    & FaST             & 44.5$\pm$0.3          & 47.1$\pm$0.1                 & 49.5$\pm$0.2 & 45.3$\pm$0.5          & 47.4$\pm$0.2                 & 49.9$\pm$0.2 \\
    & \textsc{SynthER} & 45.2$\pm$0.2 & 47.3$\pm$0.1   & ---          & 46.2$\pm$0.1 & 47.7$\pm$0.2   & ---          \\
\midrule
\multirow{2}{*}{hopper-mr}
    & FaST             & 52.9$\pm$11.5         & 87.2$\pm$2.9                 & 90.8$\pm$8.8 & 96.5$\pm$3.0          & 77.0$\pm$6.4                 & 93.8$\pm$8.0 \\
    & \textsc{SynthER} & 62.1$\pm$3.2          & 99.7$\pm$0.9    & ---          & 97.4$\pm$3.1          & 101.8$\pm$2.5   & ---          \\
\midrule
\multirow{2}{*}{walker2d-mr}
    & FaST             & 79.1$\pm$0.9          & 83.3$\pm$1.3                 & 90.7$\pm$1.0 & 77.0$\pm$1.4          & 73.6$\pm$3.3                 & 85.3$\pm$1.8 \\
    & \textsc{SynthER} & 84.7$\pm$2.5        & 90.2$\pm$0.9     & ---          & 68.6$\pm$2.5       & 82.3$\pm$4.0     & ---          \\
\bottomrule
\end{tabular}}
\caption{Each intervention of \S\ref{sec:target} applied to each generator, on the medium-replay datasets (real-data scores for these three: Table~\ref{tab:matrix}). \emph{Untargeted}: the generator's output with no targeting applied. Every arm is size-matched at $10^6$ transitions. Filter columns are the top-$20\%$ of a $5\times10^6$ pool from their own generator. Best-of-$N$ has no \textsc{SynthER} counterpart (---): a joint generator cannot natively redraw $\Pr(A \mid S)$ for a fixed state (\S\ref{subsec:related}).}
\label{tab:diffusion}
\end{table}

Best-of-10 beats filtered \textsc{SynthER} on \texttt{halfcheetah-mr}, ties or beats it on \texttt{walker2d-mr}, and trails it only on \texttt{hopper-mr}, where FaST's untargeted pool is weakest (Table~\ref{tab:main}). Unlike best-of-10, filtering needs a $5\times$ over-generation of full transitions from a generator that samples ${\sim}90\times$ slower than FaST (\S\ref{sec:exp-throughput}).

Filtering is generator-agnostic but acts differently on the two pools: it never hurts \textsc{SynthER}, whereas on FaST it rescues the weak TD3+BC pools yet degrades two of the three IQL cells, where best-of-$N$ instead gains outright (\texttt{walker2d}) or gives up far less (\texttt{hopper}).

\subsection{When advantage targeting fails}
\label{sec:exp-failure}

The medium-expert tier of Table~\ref{tab:matrix} contains every collapse in the study. On \texttt{halfcheetah-medium-expert} both targeted arms fall more than $20$ points below real data under both learners. On \texttt{hopper-medium-expert} both drop to less than half the untargeted arm under IQL but hold up under TD3+BC.

In every collapse the two mechanisms land at the same degraded level despite intervening very differently: filtering discards state-action pairs, while best-of-$N$ preserves states and redraws actions. Since both select by the same advantage estimate, the failure traces to the advantage signal itself on these datasets, not to either synthesis mechanism.

Medium-expert datasets are two-policy mixtures, and we hypothesize \emph{mode dilution}: with $V(s)$ fit to the mixture, the advantage ranking at states visited by both policies can prefer the lower-quality mode's actions, degrading the dataset exactly where imitating the expert mode was the winning strategy. For example, on IQL the collapsed halfcheetah arms land between the medium ($\sim$48) and expert ($\sim$93) levels. The failure is not universal on expert mixtures (\texttt{walker2d-medium-expert} benefits under both learners) and is learner-dependent on hopper, so dataset composition alone is an imperfect predictor. We therefore recommend advantage targeting on medium and medium-replay data.

\subsection{Inference throughput}
\label{sec:exp-throughput}

Table~\ref{tab:throughput} compares training and per-sample inference cost on a single GPU. Training is $\sim$3$\times$ slower than \textsc{SynthER} but is a one-time cost, recovered within the first few million samples at $90\times$ faster generation. The speed matters most in the over-generation regimes of \S\ref{sec:target}: a top-$20\%$ filter needs a $5\times$ pool, ${\sim}93$\,s of sampling for FaST versus ${\sim}2.4$\,h for \textsc{SynthER}, and best-of-$N$ multiplies only the action stage, a small fraction of the pipeline's per-transition cost.

\begin{table}[t]
\centering
{\footnotesize
\begin{tabular}{lccc}
\toprule
 & \textsc{SynthER} & FaST & Ratio \\
\midrule
Training (min)     & 42.5   & 128.7 & 0.33$\times$ \\
Inference / 1M (s) & 1736.0 & 18.6  & 93.3$\times$ \\
\bottomrule
\end{tabular}}
\caption{Wall-clock cost on a single RTX 2070 SUPER, averaged over the twelve datasets. Inference is the time to draw 1M transitions; \emph{ratio} is \textsc{SynthER}$\div$FaST. Per-environment numbers and per-stage breakdown in Appendix~\ref{app:throughput}.}
\label{tab:throughput}
\end{table}

\section{Discussion}

\paragraph{Offline-to-online fine-tuning.}
Fine-tuning an offline-pretrained policy online means learning from a buffer that mixes offline data with fresh interaction~\citep{nair2020awac, nakamoto2023cal, ball2023efficient}. The offline half is fixed at collection time, and what helps most is its near-on-policy share~\citep{lee2022offline}, which shrinks as the policy drifts. In robot fine-tuning, experience accrues at wall-clock speed and is punctuated by resets, so any refresh of the offline half has to fit in the gaps rather than compete with collection. Generating a million transitions costs under twenty seconds with our pipeline against roughly half an hour for \textsc{SynthER} (Table~\ref{tab:throughput}). Keeping the generator matched to a drifting policy is where our training-cost disadvantage (Table~\ref{tab:throughput}) reverses: dynamics, reward, and termination are properties of the environment rather than the policy, so those three stages stay valid, and the state model can be dropped once the live buffer supplies real states. That leaves the 1.35M-parameter action MDN as the only stage to refit, $3.2$ minutes on \texttt{hopper-medium-replay} (Appendix~\ref{app:throughput}), a cost the loop absorbs periodically rather than between rollouts. A joint model admits no such per-factor update so every refresh retrains the whole network. Advantage targeting adds nothing to either budget, since the value function enters only at candidate-scoring time (Algorithm~\ref{alg:factored}) and the agent's own improving estimate can be swapped in with no refit. Each refresh can restore the near-on-policy share that drift depletes, so the offline half adapts with the policy. We have not evaluated this regime empirically. The analysis here identifies why the factored design makes it feasible, and we leave its validation to future work.

\paragraph{Limitations.} Ancestral sampling compounds errors across five stages in a way a joint model avoids, though the transforms of \S\ref{sec:method} are designed to limit this drift and Table~\ref{tab:main} shows no resulting cost to downstream performance. Best-of-$N$ additionally shifts the action inputs of the downstream stages toward the high-advantage tail of the fitted action conditional. More fundamentally, best-of-$N$ maximizes a \emph{learned} advantage. On expert-mixture datasets targeting can fall $20$--$35$ points below real data (\S\ref{sec:exp-failure}), and the failure is environment- and learner-dependent. Our gains should therefore be read as targeting given an advantage estimate whose baseline is representative of a single, coherent behavior policy.

Because FaST generates transitions independently, it has no mechanism for multi-step coherence, so we do not evaluate the stitching-heavy domains (AntMaze, Adroit, FrankaKitchen) where trajectory-level augmentation~\citep{lee2024gta, li2024diffstitch} holds a structural advantage (Appendix~\ref{app:related-work}). Finally, we make no claims for image-based or high-dimensional sensory observations, where the state model's per-dimension, independently-parameterized conditionals (\S\ref{subsec:state}) would need to be replaced with a weight-shared, spatially-structured neural network architecture. This is left for future work.
\bibliographystyle{plainnat}
\bibliography{ref}

\clearpage
\appendix

\section{Extended related work}
\label{app:related-work}

This appendix expands on the related-work discussion of \S\ref{subsec:related} along the two axes closest to FaST's contribution: return-guided augmentation, the method most directly comparable to our pipeline, and the reweighting literature that shares our advantage signal. It closes with the density-estimation components our stages compose. A less related family of methods perturbs real transitions with Gaussian noise to regularize offline RL training~\citep{sinha2022s4rl}; these aim at robustness and overfitting control rather than distributional fidelity, so they neither model $\Pr(S, A, S', R, d)$ nor offer a mechanism to rebias coverage toward a designated region of state space.

\paragraph{Return-guided trajectory augmentation.} A parallel approach steers augmentation toward higher-return data at the \emph{trajectory} level. GTA~\citep{lee2024gta} partially noises logged trajectories and denoises them under classifier-free guidance on an amplified return. DiffStitch~\citep{li2024diffstitch} composes segments to bridge low- and high-value trajectories. These methods share our motivation (that a faithful generator reproduces mediocre behavior, so augmentation should improve dataset \emph{quality} rather than merely enlarge it) and, like our advantage targeting, act on it at generation time. Both are evaluated on the same off-policy learners we target (TD3+BC, IQL) alongside sequence models. They differ from FaST on one axis: generation granularity, with a genuine trade-off on each side. Multi-step generation enforces the trajectory-level coherence that our single-transition factorization cannot, which is why this family reports its strongest gains on sparse-reward, stitching-heavy tasks (AntMaze, Adroit, FrankaKitchen). We scope our evaluation to the transition-centric off-policy regime accordingly and leave a trajectory-level FaST to future work. In return, single-step generation reaches the action through an \emph{exact} draw from the learned $\Pr(A \mid S)$ rather than a return-conditioned distribution approximated by guidance, avoids the per-environment noise/guidance tuning these methods require, and samples at a fraction of the per-transition cost of the transition-level diffusion baseline we benchmark (\S\ref{sec:exp-throughput}).

\paragraph{Return-conditioned transition augmentation.} GODA~\citep{huang2025goda} is the closest published method to FaST in generation granularity: it augments at the level of individual transitions rather than trajectories, training a diffusion model whose samples are transition tuples and whose denoiser is MLP-based with no planning horizon, and it steers quality by conditioning on return-to-go goals drawn from the highest-return trajectories and multiplied by a scaling factor $\lambda$ ($1.1$ on D4RL) so that sampling is pushed past the returns observed in $\mathcal{D}$. The contrast with our targeting is therefore not granularity but what the intervention is able to isolate. GODA's condition applies to the tuple as a whole: states are generated jointly with actions under the goal, so selecting and raising the return target moves the generated state distribution along with the action, and the state marginal is no longer the fitted one. The scaling factor compounds this by evaluating the conditioning variable outside the range it was trained on, so the action component is an extrapolation of the learned conditional rather than an exact draw from it. Best-of-$N$ (\S\ref{sec:target}) touches the action factor alone: states come from the unmodified $\Pr(S)$, candidates are exact draws from the fitted $\Pr(A \mid S)$ at those states, and the dynamics, reward, and terminal stages are reused verbatim. The two also differ in what they require of the data, since return-to-go labels presuppose episode structure whereas our advantage scorer needs only state-action pairs.

\paragraph{Coverage and reweighting in offline RL.} A parallel line of work treats the \emph{weighting} of $\mathcal{D}$ rather than its augmentation as the lever for downstream RL quality, via advantage- or return-weighted regression~\citep{peng2019advantage, wang2018exponentially} and via coverage-aware pessimism that penalizes or excludes actions with low data support~\citep{kumar2020conservative, fujimoto2019off}. All operate on a fixed dataset without adding to it. The weighted-regression variants reweight the training objective, and retaining only a high-advantage subset of the data is the limiting case of such reweighting. Our targeted action synthesis (\S\ref{sec:target}) instead intervenes at generation time: it uses the same kind of advantage signal, but rather than up-weighting or subsetting transitions that already exist it draws new, systematically better actions from a trained generative model, producing a reusable synthetic dataset that no reweighting of $\mathcal{D}$ can express.

\paragraph{Density-estimation components.} The building blocks we employ are standard: mixture density networks~\citep{astonpr373} and autoregressive density estimation~\citep{uria2013rnade, uria2016neural, germain2015made, papamakarios2017masked}. Our state stage (\S\ref{subsec:state}) combines them with a quantile transform to form an autoregressive neural copula estimator, exploiting Sklar's decomposition of any joint distribution into marginals and a copula~\citep{joe2014dependence, nelsen2006introduction}; neural copula estimators are themselves a small but growing literature~\citep{tagasovska2019copulas, kamthe2021copula}, typically built on normalizing flows rather than autoregressive mixtures and evaluated outside offline RL.

\section{Factoring choices}
\label{app:factoring}

The chain rule of Eq.~\ref{eq:conditional} is an identity, but the particular ordering we adopt is not arbitrary. Three ordering decisions matter.

\paragraph{$\Pr(S)$ first, $\Pr(A \mid S)$ explicit.} Beginning the factorization with the marginal $\Pr(S)$ and conditioning actions on $S$ alone mirrors the behavior-policy structure of the data and is what makes targeted action synthesis (\S\ref{sec:target}) a sampling-time intervention: for a fixed synthetic state, candidate actions are redrawn from the explicit conditional and scored, while the dynamics, reward, and terminal factors are reused verbatim with no refit.

\paragraph{$R$ conditioned on $S'$.} A $\Pr(R \mid S, A)$ reward stage, the usual choice in model-based RL, is not a chain-rule factor but an assumption that $R \perp S' \mid S, A$. D4RL locomotion violates this assumption, since reward is a function of the whole transition. Dropping $S'$ would fold the dynamics noise back into the reward residual, forfeiting the near-determinism that justifies the MSE predictor of \S\ref{subsec:reward}, and would emit an $\widehat R$ drawn independently of the $\widehat{S'}$ beside it.

\paragraph{$d$ conditioned on $R$.} In D4RL locomotion, the terminal indicator is a deterministic threshold rule on $s'$ alone, so $R$ is informationally redundant. We retain it because the cost is nearly zero (the tree-based classifier of \S\ref{subsec:reward} ignores irrelevant features) and the factorization then remains valid for environments where reward and termination are coupled, as in goal-reaching tasks.

\section{Hyperparameters}
\label{app:hyperparams}

\paragraph{Generator fitting.} The state, action, reward, and delta stages are fit on $\mathcal{D}$ with early stopping on a held-out validation split. Per-stage patience values are registered in \texttt{env\_configs.py}, and the validation fraction is fixed at $0.1$ for all four stages across all environments. The terminal classifier is fit on the full $\mathcal{D}$ with no validation split or early stopping, using $50$ trees with balanced class weights and scikit-learn defaults otherwise.

\paragraph{Downstream training and reporting.} TD3+BC, IQL, and the IQL value function used for advantage scoring use the reference implementations of \citet{fujimoto2021minimalist} and \citet{kostrikov2021offline} with default hyperparameters. \textsc{SynthER} columns use the configurations released by its authors without retuning.

\paragraph{Per-stage architecture.} Per-stage architectures and their training hyperparameters are listed in Table~\ref{tab:hyperparams}. A single configuration is used throughout: the listed values apply unchanged to all twelve datasets, with no per-environment tuning. Values are registered in \texttt{env\_configs.py}.

\begin{table}[t]
\centering
\caption{Per-stage architecture and training hyperparameters. This configuration is used for every dataset in this paper. The ARMDN models each conditional with $K$ mixture components over a two-hidden-layer MLP of width $h$ (SiLU activations, LayerNorm). The action MDN places $K$ components over a residual MLP of the listed hidden widths. The delta ensemble averages $M$ independently-trained residual MLPs. The reward stage is a single residual MLP. The terminal stage is a scikit-learn \texttt{ExtraTreesClassifier} with 50 trees.}
\label{tab:hyperparams}
\footnotesize
\setlength{\tabcolsep}{3pt}
\begin{tabular}{llcccccc}
\toprule
\textbf{Stage} & \textbf{Architecture} & \textbf{Epochs} & \textbf{Batch} & \textbf{LR} & \textbf{Weight decay} & \textbf{Dropout} & \textbf{Patience} \\
\midrule
ARMDN $\Pr(S)$                  & $K=15$, $h=256$                  & 300 & 8192 & $2.8\mathrm{e}{-3}$ & $1\mathrm{e}{-4}$ & 0.10 & 30 \\
Action MDN $\Pr(A\mid S)$       & $K=15$, $(512, 512, 256)$        & 500 & 1024 & $1\mathrm{e}{-3}$   & $1\mathrm{e}{-4}$ & 0.05 & 30 \\
Delta ensemble $\Pr(S'\mid S,A)$ & $M=7$, $(512, 512, 512, 256)$   & 500 & 4096 & $2\mathrm{e}{-3}$   & $5\mathrm{e}{-4}$ & 0.04 & 25 \\
Reward MLP $\Pr(R\mid \cdot)$   & $(128, 128)$                    & 100 & 256  & $3\mathrm{e}{-4}$   & $1\mathrm{e}{-4}$ & 0.05 & 15 \\
Terminal $\Pr(d\mid \cdot)$     & ExtraTrees $\times 50$, balanced & ---  & ---   & ---                  & ---                & ---   & --- \\
\bottomrule
\end{tabular}
\end{table}

\section{Per-stage training and inference cost}
\label{app:throughput}

All measurements in Table~\ref{tab:throughput} use a single NVIDIA GeForce RTX 2070 SUPER (8 GB) with fp32 throughout. Table~\ref{tab:stage-cost} breaks down the wall-clock cost of FaST by stage on \texttt{hopper-medium-replay-v2}. The delta ensemble dominates both training and per-sample inference. Table~\ref{tab:throughput-app} gives the per-environment breakdown of Table~\ref{tab:throughput}: each environment row is averaged across its three D4RL variants (medium, medium-replay, medium-expert for locomotion; umaze, medium, large for maze2d), and inference is the time to draw 1M transitions (mean $\pm$ std, 3 repeats).

\begin{table}[t]
\centering
\caption{Per-stage cost on \texttt{hopper-medium-replay-v2}. Params counts the trained parameters of the stage. Sampling cost is microseconds per transition. No stage is refit for advantage targeting (Algorithm~\ref{alg:factored}). Best-of-$N$ over-generates only from the fitted action stage, whose sampling cost is a small fraction of the pipeline's.}
\label{tab:stage-cost}
\footnotesize
\setlength{\tabcolsep}{5pt}
\begin{tabular}{llccc}
\toprule
\textbf{Stage} & \textbf{Architecture} & \textbf{Params} & \textbf{Train (min)} & \textbf{Sample ($\mu$s/trans.)} \\
\midrule
$\Pr(S)$          & quantile--logit ARMDN & 0.80M & 7.25  & 2.71 \\
$\Pr(A\mid S)$    & logit MDN             & 1.35M & 3.15  & 0.79 \\
$\Pr(S'\mid S,A)$ & delta ensemble $\times 7$ & 13.0M & 21.78 & 6.93 \\
$\Pr(R\mid \cdot)$ & residual MLP             & 70.3K & 6.79  & 0.23 \\
$\Pr(d\mid \cdot)$ & ExtraTrees $\times 50$   & 12.2K nodes & 0.04  & 0.41 \\
\bottomrule
\end{tabular}
\end{table}

\begin{table}[t]
\centering
\caption{Training and inference wall-clock cost on a single RTX 2070 SUPER. Each row is averaged over the three D4RL variants of that environment: medium, medium-replay, medium-expert for locomotion and umaze, medium, large for maze2d. Inference is the time to draw 1M transitions (mean $\pm$ std over 3 repeats). \emph{Speedup} is the \textsc{SynthER}-to-FaST inference ratio.}
\label{tab:throughput-app}
\footnotesize
\setlength{\tabcolsep}{4pt}
\begin{tabular}{lccccc}
\toprule
 & \multicolumn{2}{c}{\textbf{Training (min)}} & \multicolumn{2}{c}{\textbf{Inference / 1M (s)}} & \\
\cmidrule(lr){2-3} \cmidrule(lr){4-5}
\textbf{Environment} & \textsc{SynthER} & FaST & \textsc{SynthER} & FaST & \textbf{Speedup} \\
\midrule
halfcheetah-v2 & 40.2 & 111.4 & 1735.08 $\pm$ 1.25 & 18.98 $\pm$ 7.27 & 91.4$\times$ \\
hopper-v2      & 40.0 & 110.1 & 1732.73 $\pm$ 2.17 & 16.06 $\pm$ 5.61 & 107.9$\times$ \\
walker2d-v2    & 40.3 & 181.3 & 1740.54 $\pm$ 2.06 & 18.74 $\pm$ 7.39 & 92.9$\times$ \\
maze2d-v1      & 49.4 & 111.9 & 1735.82 $\pm$ 0.18 & 20.66 $\pm$ 9.97 & 84.0$\times$ \\
\midrule
\textbf{Mean}  & \textbf{42.5} & \textbf{128.7} & \textbf{1736.04} & \textbf{18.61} & \textbf{93.3$\times$} \\
\bottomrule
\end{tabular}
\end{table}

\section{State model diagnostics}
\label{app:ablations}

\begin{figure}[t]
    \centering
    \includegraphics[width=\textwidth]{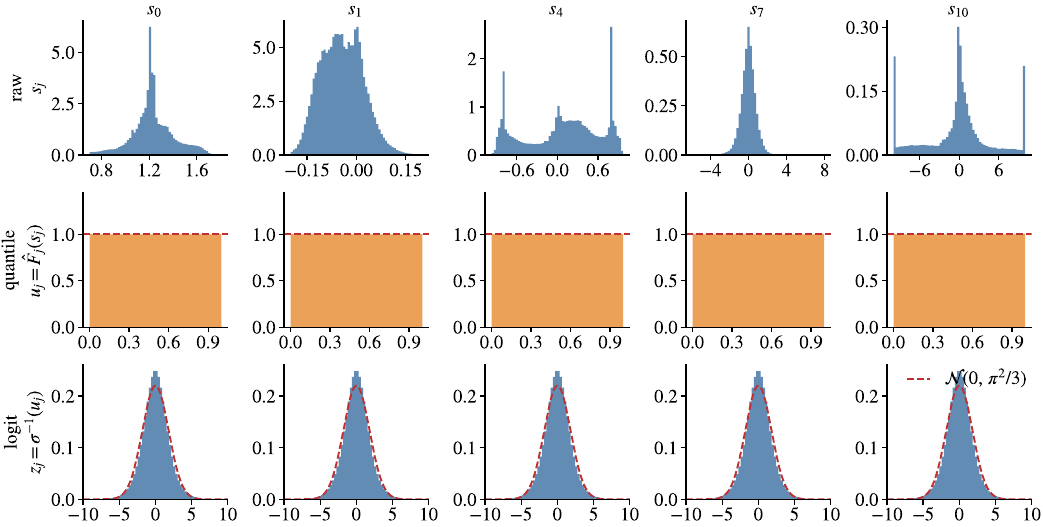}
    \caption{Quantile-logit transform applied per-dimension to state dimensions $0, 1, 4, 7, 10$ of \texttt{hopper-medium-replay-v2}. Top: raw marginals, multimodal and heavy-tailed with spikes at bounded extremes. Middle: after the empirical CDF $\hat F_j$, yielding $\mathrm{Uniform}(0,1)$ by construction. Bottom: after the logit, yielding a standard logistic distribution and the dashed overlay shows the close Gaussian approximation that the MDN's components exploit.}
    \label{fig:state}
\end{figure}

This appendix collects supporting material for \S\ref{subsec:state}: implementation details of the quantile--logit transform and the rationale for strict in-support sampling. Figure~\ref{fig:state} shows the effect of the bijection of Eq.~\ref{eq:qlogit-fwd} on five state dimensions of \texttt{hopper-medium-replay-v2}.

\paragraph{Why the empirical CDF and not a smoothed estimate.} We estimate $\hat F_j$ by ranks rather than by integrating a kernel-density (KDE) estimate of the marginal. A kernel CDF has unbounded support, so its inverse would place synthetic mass outside the observed range $[\min_i s_{i,j},\,\max_i s_{i,j}]$, forfeiting the strict in-support guarantee of \S\ref{subsec:state} (property (i)). Smoothing would also blur the sharp mass concentrations at physical joint limits (Figure~\ref{fig:state}, row 1). The downstream MDN fits such a spike with a narrow mixture component, and the inverse CDF $\hat F_j^{-1}$ restores the exact tied value.

\paragraph{In-support sampling rationale.} Strict in-support sampling avoids synthesizing physically invalid configurations such as joint limits and implausible velocities. It is also crucial that downstream factors don't see these invalid configurations since they were trained on real states. The tradeoff is that we forgo explicit extrapolation beyond $\mathcal{D}$.

\section{Action model diagnostics}
\label{app:action-details}

\begin{figure}[t]
    \centering
    \includegraphics[width=0.6\linewidth]{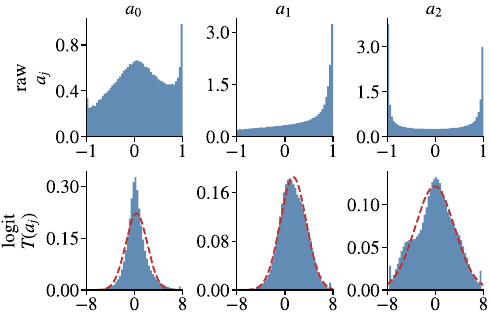}
    \caption{Action marginals on dimensions $0, 1, 2$ of \texttt{hopper-medium-replay-v2},
before (top) and after (bottom) the logit transform. Raw actions pile up
at the actuator limits $\pm 1$. Under $T$, the boundary saturation is
remapped to finite locations on $\mathbb{R}$ and the resulting distribution
is well-approximated by a single Gaussian (red dashed overlay), giving
the MDN a target that Gaussian components can fit.
}
    \label{fig:action}
\end{figure}

This appendix collects supporting material for \S\ref{subsec:action}: contrast with the quantile--logit transform and the rationale for $K=15$ mixture components. Figure~\ref{fig:action} shows the effect of the bijection of Eq.~\ref{eq:action-logit} on three action dimensions of \texttt{hopper-medium-replay-v2}.

\paragraph{Contrast with the quantile--logit transform of \S\ref{subsec:state}.} $T$ is a fixed function, not estimated from data. The quantile step of \S\ref{subsec:state} serves to separate marginal shape from dependence, which is worth its cost when marginals are multimodal and heavy-tailed. Action marginals are neither: their awkward feature is the pile-up at $\pm 1$, and handling that bounded support is all that $T$ needs to do.

\paragraph{Mixture-component count.} Table~\ref{tab:action-k-sweep} sweeps the number of mixture components $K \in \{1, 3, 5, 8, 15\}$ on \texttt{hopper-medium-replay-v2} (3 seeds each), reporting held-out NLL of the logit-transformed action and the KS statistic between synthetic and real action marginals. Because the action conditional is low-dimensional ($d = 3$) and the logit transform of Eq.~\ref{eq:action-logit} renders each marginal near-Gaussian, fit quality saturates by a handful of components: NLL improves by only $0.05$ nats from $K=1$ to $K=15$, is converged by $K \approx 3$ within seed noise, and the marginal-fidelity KS shows no trend across the range. We therefore fix $K=15$ generously (Table~\ref{tab:hyperparams}) rather than tuning it per dataset.

\begin{table}[h]
\centering
\caption{Action-model component sweep on \texttt{hopper-medium-replay-v2}, mean $\pm$ std over 3 seeds. NLL is the negative log-likelihood of the logit-transformed action on a held-out split (lower is better). KS is the Kolmogorov--Smirnov statistic between synthetic and real action marginals.}
\label{tab:action-k-sweep}
\footnotesize
\begin{tabular}{ccc}
\toprule
\textbf{$K$} & \textbf{NLL} & \textbf{KS} \\
\midrule
1 & 4.636 $\pm$ 0.014 & 0.0151 $\pm$ 0.0041 \\
3 & 4.589 $\pm$ 0.013 & 0.0201 $\pm$ 0.0022 \\
5 & 4.586 $\pm$ 0.013 & 0.0123 $\pm$ 0.0005 \\
8 & 4.577 $\pm$ 0.001 & 0.0153 $\pm$ 0.0022 \\
15 & 4.583 $\pm$ 0.003 & 0.0148 $\pm$ 0.0012 \\
\bottomrule
\end{tabular}
\end{table}

\section{Next-state model details}
\label{app:delta-diagnostics}

This appendix provides quantitative evidence for two design choices in \S\ref{subsec:next} (next-state model): (i) the unimodal-conditional assumption underlying MSE regression, and (ii) ensemble size. All results are on \texttt{hopper-medium-replay-v2}.

\begin{figure}[t]
\centering
\includegraphics[width=\textwidth]{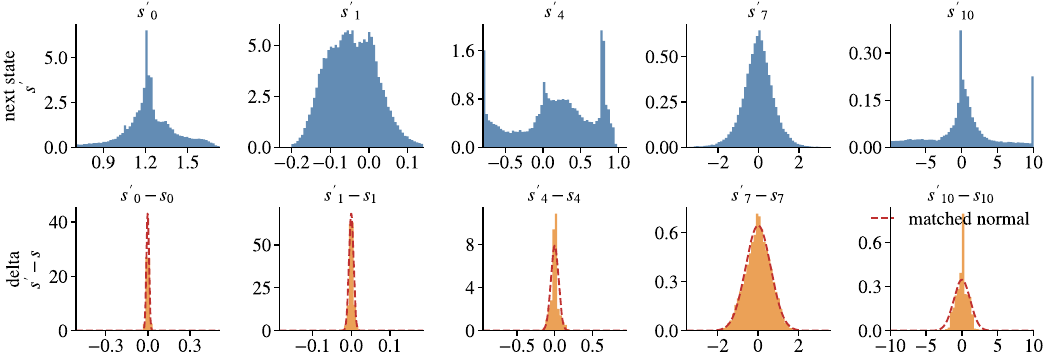}
\caption{Next-state marginals (top) against delta marginals $\delta_j = s'_j - s_j$ (bottom) for dimensions $0, 1, 4, 7, 10$ of \texttt{hopper-medium-replay-v2}, on a set of non-terminal transitions. Raw next-state marginals are broad and multimodal with mass piled at joint and velocity limits (dim $4$), while the corresponding deltas are unimodal and concentrated at zero. The red dashed curve is a normal matched to each panel's mean and variance, drawn as a fixed reference for width and shape only. The model does not assume normality.}
\label{fig:delta-marginals}
\end{figure}

\paragraph{Shape of the residuals.}
Figure~\ref{fig:delta-marginals} shows what the delta parameterization buys. Predicting $S'$ directly means fitting marginals that are broad, multimodal, and spiked at physical limits. Subtracting $S$ collapses the position and angle coordinates onto a single tight mode at zero, which is the shape an MSE regressor is well matched to. We do not claim the residuals are Gaussian, and \S\ref{subsec:next} does not require it. A point predictor needs only a unimodal distribution, so that the conditional mean is a representative summary rather than an average over separated modes.

\paragraph{Ensemble size ablation.}
Table~\ref{tab:delta-ensemble-size} sweeps the number of ensemble members $M \in \{1, 3, 5, 7\}$ on held-out test-set R$^2$ and RMSE over 3 seeds. R$^2_\delta$ plateaus around $M=5$, with a further gain of only $0.005$ from $M=5$ to $M=7$. The single-regressor baseline ($M=1$) is clearly worse (R$^2_\delta$ $=0.830$ vs.\ $0.877$ at $M=7$), so ensemble averaging does reduce variance. We use $M=7$ because the extra cost is modest and it is the most consistent across random seeds.

\begin{table}[h]
\centering
\caption{Ensemble size sweep on \texttt{hopper-medium-replay-v2}. R$^2_\delta$ measures prediction accuracy on the delta target. RMSE in next-obs space. Mean over 3 seeds.}
\label{tab:delta-ensemble-size}
\footnotesize
\setlength{\tabcolsep}{5pt}
\begin{tabular}{cccc}
\toprule
\textbf{$M$} & \textbf{R$^2_{\text{next-obs}}$} & \textbf{R$^2_{\delta}$} & \textbf{RMSE} \\
\midrule
1 & 0.9931 $\pm$ 0.0004 & 0.8299 $\pm$ 0.0144 & 0.1184 $\pm$ 0.0046 \\
3 & 0.9945 $\pm$ 0.0001 & 0.8615 $\pm$ 0.0039 & 0.1027 $\pm$ 0.0026 \\
5 & 0.9950 $\pm$ 0.0002 & 0.8724 $\pm$ 0.0055 & 0.0985 $\pm$ 0.0032 \\
7 & 0.9951 $\pm$ 0.0004 & 0.8769 $\pm$ 0.0083 & 0.0968 $\pm$ 0.0029 \\
\bottomrule
\end{tabular}
\end{table}

\paragraph{Terminal transitions.}
Terminal transitions pair a pre-fall state with a simulator-reset next-state. Including them would corrupt the conditional mean, so they are excluded from delta-ensemble training.

\section{Reward model diagnostics}
\label{app:reward-diagnostics}

This appendix provides quantitative evidence on the reward conditional's shape to justify the choice of MSE over a mixture model.

\begin{figure}[t]
\centering
\includegraphics[width=\textwidth]{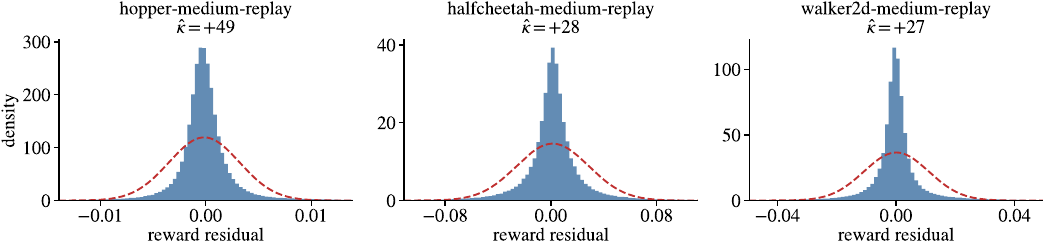}
\caption{Reward residuals $r - f_{\theta_R}(s, a, s')$ on non-terminal transitions of each locomotion environment. All three distributions are sharply peaked and unimodal, which is all the MSE choice of \S\ref{subsec:reward} requires. The matched Gaussian overlay (red) badly under-covers the central spike, confirming the conditional is unimodal but non-Gaussian. The excess kurtosis $\hat\kappa$ above each panel measures the other half of that misfit, the heavy tails, which lie outside the plotted window. The maze2d datasets are omitted because their reward is binary; they are treated separately below.}
\label{fig:reward-residuals}
\end{figure}

\paragraph{Residual shape: why MSE suffices.}
The state and action stages use mixture density networks because their conditionals
are multimodal (\S\ref{subsec:state}, \S\ref{subsec:action}). The reward stage faces
the same question: is the conditional multimodal? Figure~\ref{fig:reward-residuals}
shows it is not. Residuals are sharply peaked and unimodal across all three
locomotion environments.

\paragraph{Binary rewards on maze2d.}
Rewards in maze2d are sparse and binary, taking value $1$ when the goal is reached and $0$ otherwise. Our reward model outputs a continuous value, which we binarize with a rank-based cut: we assign $1$ to the highest-scoring transitions and $0$ to the rest, choosing the cutoff so that the fraction of positive rewards matches that of the source dataset. Though a classifier is the principled model for such a target, we use a regression head for consistency with the rest of the pipeline and leave classification-based reward models to future work.

\section{Advantage-targeted generation: details and design choices}
\label{app:target-details}

\paragraph{Advantage estimator.}
We fit IQL~\citep{kostrikov2021offline} on the real dataset $\mathcal{D}$ with the reference implementation and default hyperparameters, and score state-action pairs with the advantage $A_{\mathrm{IQL}}(s,a) = \min(Q_1, Q_2)(s,a) - V(s)$ of \S\ref{subsec:background}. The value function sees exactly the data the generator sees: no additional environment interaction and no held-out data.

\paragraph{Filter definition.}
The enrichment baseline generates a pool of $5\times$ the target dataset size, scores every transition with $A_{\mathrm{IQL}}(\widehat S, \widehat A)$, and keeps the top $20\%$ (a hard threshold at the pool's 80th percentile), so the filtered dataset is size-matched to the untargeted one at $10^6$ transitions and selection is not confounded with dataset size.

\begin{figure}[t]
\centering
\includegraphics[width=0.6\linewidth]{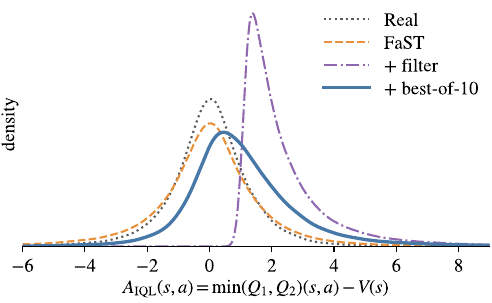}
\caption{Advantage of the transitions from multiple sources on \texttt{hopper-medium-replay-v2}. The untargeted FaST pool tracks the real data, so fidelity alone cannot raise action quality. Best-of-10 shifts the bulk right (mean $+1.15$) while keeping the low-advantage tail. The top-$20\%$ filter reaches $+2.53$ instead by truncating it (10th percentile $+1.23$, against $-0.77$).}
\label{fig:advantage-wall}
\end{figure}

\paragraph{Dataset regimes.}
The targeting evaluation covers all nine locomotion datasets (Table~\ref{tab:matrix}), and the results validate a headroom account of when targeting helps. Medium and medium-replay logs come from a single mediocre policy or a learning policy and therefore contain a spread of action quality at overlapping states. This is the signal the advantage scorer exploits, for filtering and best-of-$N$ alike, and both mechanisms gain on nearly all of these datasets. Medium-expert logs are two-policy mixtures, meaning the value baseline is fit to a bimodal return distribution and targeting can be actively harmful (\S\ref{sec:exp-failure}), failing identically for both mechanisms. Whether to apply targeting is therefore a question of dataset composition, not of advantage-scorer applicability. The three maze2d datasets of Table~\ref{tab:main} are excluded from the targeting evaluation. Their reward is sparse and binary (Appendix~\ref{app:reward-diagnostics}), so away from the goal region no candidate action collects reward and $A_{\mathrm{IQL}}$ differences come entirely from bootstrapped value estimates. Whether that signal is strong enough to rank candidates for best-of-$N$ is an empirical question we did not address in this work. We scoped the evaluation to the dense-reward locomotion suite.

\paragraph{Best-of-$N$ dose-response.} Table~\ref{tab:bon-sweep-app} sweeps the candidate count on \texttt{walker2d-medium-replay-v2} under TD3+BC, referenced from \S\ref{sec:exp-target}. The gain is monotone but sharply saturating: $79.1$ at $N{=}1$ (the untargeted pool) rises to $90.0$ at $N{=}5$. Past that point the increments are small, $+0.7$ from $N{=}5$ to $N{=}10$ ($p = 0.37$) and $+1.2$ from $N{=}5$ to $N{=}20$ ($p = 0.05$). $N{=}5$ is therefore the cost-effective setting, and the $N{=}10$ used throughout the experiments is a conservative rather than a tuned choice.

\begin{table}[h]
\centering
\caption{Best-of-$N$ dose-response on \texttt{walker2d-medium-replay-v2} (TD3+BC). $N=1$ is the untargeted pool. Scores are means over the final $500$k steps, $3$ generation $\times\ 3$ training seeds, $\pm$ std across the generation-seed means, as in the uniform protocol.}
\label{tab:bon-sweep-app}
\small
\begin{tabular}{l c c c c}
\toprule
$N$ & $1$ (pool) & $5$ & $10$ & $20$ \\
\midrule
Score & $79.1 \pm 0.9$ & $90.0 \pm 0.6$ & $90.7 \pm 1.0$ & $91.2 \pm 1.1$ \\
\bottomrule
\end{tabular}
\end{table}

\paragraph{Interaction between the two mechanisms.}
Figure~\ref{fig:advantage-wall} shows what each mechanism of \S\ref{sec:target} does to the advantage distribution, and the two are not interchangeable: best-of-$N$ shifts it while filtering truncates it. They also do not stack. Applying the top-$20\%$ advantage filter to a best-of-10 pool on \texttt{walker2d-medium-replay-v2} fails to improve on best-of-10 alone under TD3+BC: $-2.0 \pm 2.5$ D4RL points. Once the actions are optimized per state, filtering can only act by discarding states, and whatever advantage it adds does not survive that cost. We ran this experiment on one dataset under one algorithm.

\end{document}